\documentclass[10pt,twocolumn,letterpaper]{article}

\usepackage{cvpr}
\usepackage{times}
\usepackage{epsfig}
\usepackage{graphicx}
\usepackage{amsmath}
\usepackage{amssymb}
\usepackage{algorithm}
\usepackage[noend]{algpseudocode}
\usepackage{authblk}

\usepackage[pagebackref=true,breaklinks=true,letterpaper=true,colorlinks,bookmarks=false]{hyperref}

\cvprfinalcopy 

\def\cvprPaperID{500} 

\ifcvprfinal\fi
\begin{document}

\title{Generating Synthetic X-ray Images of a Person from the Surface Geometry}

\author[1]{Brian Teixeira}
\author[1]{Vivek Singh}
\author[1]{Terrence Chen}
\author[1]{Kai Ma}
\author[1]{Birgi Tamersoy}
\author[2]{Yifan Wu}
\author[3]{Elena Balashova}
\author[1]{Dorin Comaniciu}
\affil[1]{Medical Imaging Technologies, Siemens Healthineers, Princeton, NJ, USA}
\affil[2]{Temple University, Philadelphia, PA, USA}
\affil[3]{Princeton University, Princeton, NJ, USA}

\maketitle
\thispagestyle{empty}

\begin{abstract}
We present a novel framework that learns to predict human anatomy from
body surface. Specifically, our approach generates a synthetic X-ray
image of a person only from the person's surface geometry.
Furthermore, the synthetic X-ray image is parametrized and can be
manipulated by adjusting a set of body markers which are also
generated during the X-ray image prediction. With the proposed
framework, multiple synthetic X-ray images can easily be generated by
varying surface geometry. By perturbing the parameters, several additional synthetic
X-ray images can be generated from the same surface geometry. As a result, our approach offers a
potential to overcome the training data barrier in the medical
domain. This capability is achieved by learning a pair of networks -
one learns to generate the full image from the partial image and a set
of parameters, and the other learns to estimate the parameters given
the full image. During training, the two networks are trained
iteratively such that they would converge to a solution where the
predicted parameters and the full image are consistent with each
other. In addition to medical data enrichment, our framework can also
be used for image completion as well as anomaly detection. \footnote{This feature is based on research, and is not commercially available. Due to regulatory reasons its future availability cannot be guaranteed.}
\end{abstract}

\section{Introduction}
\begin{figure}[t]
	\includegraphics[width=\linewidth]{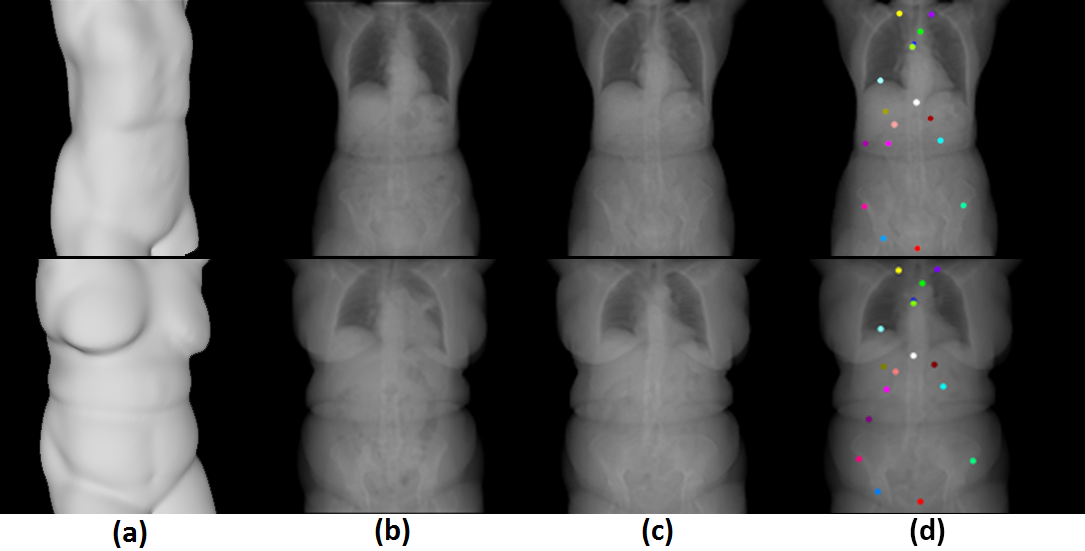}
	\caption{Synthetic X-ray generated from 3D surface meshes.
          Each row contains data corresponding to one patient, (a) 3D
          surface meshes (b) corresponding ground truth images
          obtained using medical scanning, (c) predicted X-ray images
          using the proposed approach, and (d) the X-ray images with
          predicted markers.}
	\label{fig:intro}
\end{figure}

Over the past decade, there has been significant advances in realistic
human body shape modeling and simulation in the graphics
domain \cite{SCAPE, SMPL, breath, DYNA}, where different statistical
models have been applied to learn compact parametric representations of
the human body shape. However, their impact on the
healthcare domain is relatively limited \cite{DARWIN, shapeadip}. One
major reason is that existing shape modeling approaches
focus primarily on the skin surface while the healthcare domain pays
more attention to the internal organs. This work attempts
to tackle this limitation by addressing the challenging task of
estimating the internal anatomy of human body from the surface
data. More specifically, our approach generates a synthetic X-ray
image of a person only from the surface geometry. However the
synthetic X-ray would only serve as an approximation of the true
internal anatomy, we thus simultaneously predict markers which can be
adjusted to update the X-ray image. Since the markers serve as spatial
parameters that can be used to perturb the image, we refer to
the predicted image as a \textit{parametrized image}. Figure
\ref{fig:intro} shows examples of parametrized X-ray images generated
from surface data.

Learning to predict parametrized images is a very challenging task due
to strict constraints in the output space. As the training framework
learns to predict images and the corresponding spatial
parameters (i.e. markers), it also needs to ensure that the
perturbations of these parameters lies on a manifold of
'realistic deformations' (e.g. realistic facial
expressions when generating face images or realistic body anatomy when
generating synthetic X-ray). Since learning such output spaces (which
are implicitly highly correlated) is difficult, we propose to learn a
pair of networks, one trained to predict the parameters from image
contents, and the other trained to predict the image contents from the
parameters. When the parameters are updated, the networks are applied
iteratively in a loop until convergence. To facilitate such convergent
behavior during test phase, we present a novel learning algorithm that
jointly learns both the networks. While some recent work have utilized
predicting markers as a supplementary task within the context of
multi-task learning \cite{guided, face_land}, to the best of our
knowledge, this work is first in explicitly learning a bijection
between the predicted markers and the generated images.

We use the proposed framework to generate synthetic X-ray image from 3D body surface meshes. 
Such a technology can be used in conjunction with existing approaches to estimate 3D body surface models from depth \cite{DARWIN} and potentially benefit medical procedures such as patient positioning for scanning or interventional procedures.
We report several impactful
applications of this technology such as anomaly detection as well as
completion of full X-ray from partial X-ray. We also demonstrate via
experiments that the parametrized X-ray images can be used to generate
training data thus helping to overcome the significant big data
barrier faced during the application of the deep learning approaches
in medical imaging tasks.

The key contributions of this paper can be summarized as follows:
\begin{itemize}
\item A novel framework to generate parametrized images using a
  convergent training pipeline.
\item A novel technology to predict parametrized X-ray images from body
  surface data, which as we demonstrate can significantly impact the
  healthcare domain.
\item Use of Wasserstein GANs (WGAN) \cite{wgan} with gradient penalty
  method for a conditional regression task. 
\end{itemize}

\section{Related Work}
Learning parametrized image representations can be formulated as
\textbf{multi-task learning}, where one task is to generate the synthetic image
from input data, and other is to ensure that the synthetic image
correlates strongly with the set of control parameters. \cite{guided}
presented such a multi-task architecture to regress both facial
markers and a corresponding 3D face represented as a volumetric
image. They present a network architecture with two stacked hourglass
networks \cite{hourglass}, one predicts the markers, and the other
predicts the volumetric images taking output of the marker network as
input. While they show the use of landmarks improve face prediction,
there is no guarantee that the predicted markers and generated face
converge to have a strong correlation, as the markers are pre-trained
independently from the face network and never get any feedback from it. \cite{hyperface} employs an architecture where tasks are
incorporated with a single network but these auxiliary tasks are
categorical such as gender recognition. While such networks would have
the ability to implicitly learn the correlation between the two tasks
(i.e. parameter space and image space in our setting), they are
difficult to train because of the large variation in losses during the
training period \cite{face_land}.

\begin{figure*}
	\centering
        \includegraphics[width=0.8\linewidth]{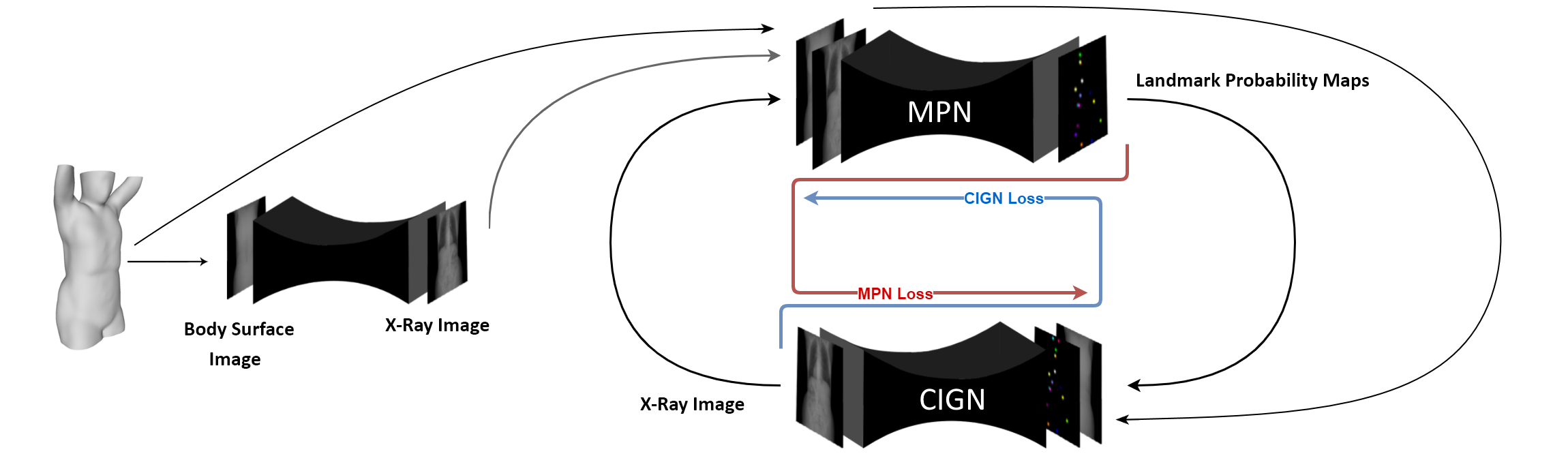}
	\caption{Pipeline to generate Parametrized Images. First, the
          synthetic X-ray image is generated from the surface data to
          provide an initial estimate. Next, the landmarks are
          detected in X-ray image, which are then in turn re-utilized
          to predict the X-ray image. After a few iterations, both the
          X-ray image and marker distribution converge, and become
          tightly correlated.}
	\label{fig:overview}
\end{figure*}

With the advent of generative adversarial networks \cite{gan}, several
attempts have been made to generate synthetic training data or augment
data in domain where it's either difficult to acquire or annotations
are difficult to obtain. Although \cite{rendergan} focuses on
generating data from noise, it bears similarities with our approach as
it incorporates semantic parametrization during the image generation
and shows that generated data can help with training deep
networks. However, the sampled parameter space is limited to global
scene parameters such as lighting variation, background variation
etc., while in our work, we focus on the spatial parametrization which
allow changing the scene structure. \cite{zhu2016generative} employs
GANs to learn the natural image manifold and allow user to manipulate
the generated image by making structural edits, which are provided as
sketches and image is updated by finding the nearest instance on the
manifold. In contrast, our approach learns an explicit mapping between
the image and several markers, and provides editability using these
markers. \cite{cyclegan} presents an approach for image domain
transfer by learning a pair of networks with an inverse relationship,
using cyclic consistency as a regularizer. However our approach learns
mapping from the source domain (surface data) to target domain (X-ray
image) together with control parameters (markers), and consistency is
being enforced between the target image space and parameter space.

\section{Parametrized Images}
We refer \textit{parametrized image} to an image that is parametrized
by a set of spatially distributed markers, which can be perturbed to
manipulate the contents of the image to generate realistic image
variations. Such manipulation of an image via markers (spatial
parameters) requires learning a bijection mapping. In this work, we
achieve this by learning a pair of networks, \textit{Marker Prediction
  network} (MPN), trained to predict the parameters from the image
contents, and \textit{Conditional Image Generation network} (CIGN)
trained to predict the image contents given the parametrization. We
use the networks to predict an initial estimate and then iteratively
refine until convergence. Figure \ref{fig:overview} shows an overview
of the networks involved in the parametrized image generation.  While
parametrized images can be generated from noise (similar to image
generation task \cite{gan}), we focus on the task of conditional image
generation \cite{pixpix} as it naturally applies to task of generating
the X-ray images from 3D body surface data.

We represent the 3D human surface mesh data with a 2-channel 2D image;
the first channel stores the depth of the body surface as observed
from front, and second channel stores the thickness computed by
measuring the distance between the closest and furthest point as
observed from front; in the rest of the document we refer to this 2
channel image as \textit{surface image}.

\subsection{Marker Prediction Network}
The marker prediction network takes the surface image as well as the
predicted X-ray image as input and predicts the locations for all the
markers. We employ a U-Net like \cite{unet} architecture to train a
network to regress from a 3-channel input image (2 surface data
channels, 1 X-ray image channel) to a 17-channel heatmap image by minimizing L2-loss;
these heatmaps correspond to anatomically meaningful landmarks
(such as lung top, liver top, kidney center etc). Each output channel
compares with the given ground truth that includes a Gaussian mask
(kernel $radius=5$, $\sigma=1$) centered at the given target location
( similar to other landmark detection approaches
\cite{fashion_land,surgery_land,robust_land}). 

\subsection{Conditional Image Generation Network}
The proposed conditional image generation network is derived from the
conditional GAN architecture \cite{pixpix}. The generator with the
U-Net architecture takes the surface image and marker heatmaps as
input and outputs the synthetic X-ray image. To stabilize the
CIGN training, we adopt the Wasserstein loss with gradient
penalty introduced in \cite{iwgan}, which is known to outperform other
adversarial losses. The critic takes the surface image and
corresponding X-ray image as input, though theoretically, a better
critic model would have taken the surface images, marker maps
(parameter space) and X-ray image as input to implicitly force a
strong correlation between them; however fusing all the data together
as a $20$ channel image did not help training a useful critic in
any of the GAN variants \cite{pixpix, wgan, iwgan}; we noticed that
the critic was not able to utilize the marker maps to determine the
real vs fake, and thus fails to provide gradients that would eventually
help generating the X-ray image. The final choice of the network
architecture was determined after thorough experimentation with
several architectures as well as GAN variants.

\section{Learning Parametrized Image Representation}
In this section, we describe the procedure to train the networks
(marker prediction as well as conditional image generation
network). We first pre-train the networks using the available ground
truth data, and subsequently refine them end-to-end to minimize the
combined loss, defined as,
\begin{equation}
\mathcal{L} = \mathcal{L}_{MPN} + \mathcal{L}_{CIGN}
\end{equation}
where, $\mathcal{L}_{MPN}$ is the mean squared error between the
predicted and the ground truth heat maps for the markers,
$\mathcal{L}_{CIGN}$ is loss between the predicted and ground
truth X-ray image.

\subsection{Pre-training Marker Prediction Network}
We pre-train the marker network using the Adam optimizer
\cite{kingma2015} to minimize the MSE loss, and set the initial
learning rate to $10^{-3}$. During pre-training, we use the ground
truth X-ray images with body surface images as input.

During the convergent training process, the input is replaced by the
predicted X-ray image. This initially worsens the performance on the
marker prediction network but it quickly recovers after a few epochs of
convergent training, as demonstrated in the experiments.

\subsection{Pre-training Conditional Image Generation Network}
We pre-train the image generation network with surface images and
ground truth landmark maps as input, using the Adam optimizer with
initial learning rate of $10^{-3}$. After pre-training, we use the
RMSProp \cite{rmsprop} with a low learning rate of $10^{-5}$. In our
experiments, we found the gradient penalty variant of WGAN
\cite{iwgan} to outperform the original WGAN with
weight clipping \cite{wgan}. The architecture of the critic was
similar to the encoder section of the generator network. We observe
that in case of WGAN, using a more complex critic helps converging faster.

During the convergent training, the network is iteratively updated
using the predicted landmarks as input.

\subsection{Convergent Training via Iterative Feedback}
During the test phase, we apply both networks iteratively in
succession until both of them reach the steady state. This implicitly
requires the networks to have a high likelihood of convergence during
the training stage. A stable solution sits where both the markers/parameters
and synthetic image are in complete agreement with each other,
suggesting a bijection. We achieve the goal by freezing one network and
updating the weights of the other network using its own loss as well
as the loss backpropagated from the other network. Thus, not only the
networks get feedback from the ground truth, they also get feedback on
how they helped each other (good markers give good X-ray image, and
vice versa). The gradient flow during backpropagation is
also shown in Figure \ref{fig:overview}.

The losses optimized by conditional image generation (\textit{CIGN}) and
marker prediction network (\textit{MPN}) at each iteration are given by,
\begin{eqnarray}
\mathcal{L_{CIGN}} = \mathcal{L}_{adv}(I_{gt}, I^i_{syn}) +
\mathcal{L}_{2}(MPN(I^i_{syn}, S), M_{gt}) \\ \mathcal{L_{MPN}} =
\mathcal{L}_{2}(M^i, M_{gt}) + \mathcal{L}_{1}(I_{gt}, CIGN(M^i, S))
\end{eqnarray}
where, $CIGN(.)$ and $MPN(.)$ are deep networks depicted in functional
form; $I_{gt}$ and $M_{gt}$ are ground truth image and markers heat maps
respectively; $I^i_{syn}$ and $M^i$ are predicted images and markers
heat maps at iteration $i$.

The iterative approach to train the networks to facilitate convergence
is motivated by the iterative adversarial training procedure
explained in \cite{gan}. However, in this case, the
networks are learning to cooperate instead of compete. Similar to GAN
training, there is a possibility that the training may become unstable
and diverge. We address this issue by weighting the losses with
appropriate scale.

While the number of epochs required to reach convergence depends on
how tightly the output of the two networks correlate, in our
experiments, we found $50$ epochs to be sufficient. After these
epochs, no significant change in X-ray or in landmarks positions were
observed, suggesting convergence. Algorithm \autoref{algo} details the
pseudo-code for convergent training.

\begin{algorithm}[t!]
  \caption{Convergent training}\label{algo}
  \begin{algorithmic}[1]
    \Require: $\alpha$, the learning rate. $T_s$, the model generating
    X-ray from surface. $w_m$, initial MPN
    weights. $w_t$, initial CIGN weights. $\lambda_1$
    and $\lambda_2$ loss scaling factors.
    \While{$w_m$ and $w_t$ have not converged}
    \State Sample $x$ a batch from the input data.
    \State Get $\hat{m}, \hat{t}$ ground truth landmarks and X-ray
    \State $t_1 \gets T_s(x) $
    \State $x' \gets concat(x, t_1)$
    \State $m_1 \gets MPN(x')$
    \State $e_M \gets \mathcal{L_{MPN}}(m_1, \hat{m})$
    \State $x' \gets concat(x, m_1)$
    \State $t_2 \gets CIGN(x')$
    \State $e_T \gets \mathcal{L_{CIGN}}(t_2, \hat{t})$
    \State $g_{w_m} \gets \nabla_{w_m} \left[e_M + \lambda_1 \cdot e_T \right]$
    \State $w_m \gets w_m + \alpha \cdot \text{RMSProp}(w_m, g_{w_m}) $

    \State $x' \gets concat(x, m_2)$
    \State $m_2 \gets MPN(x')$
    \State $e_M \gets \mathcal{L_{CIGN}}(m_2, \hat{m})$
    \State $g_{w_t} \gets \nabla_{w_t} \left[e_T + \lambda_2 \cdot e_M \right]$
    \State $w_t \gets w_t + \alpha \cdot \text{RMSProp}(w_t, g_{w_t}) $
    \EndWhile
\end{algorithmic}
\end{algorithm}

\begin{figure*}[]
  \begin{center}
    \includegraphics[width=0.8\linewidth]{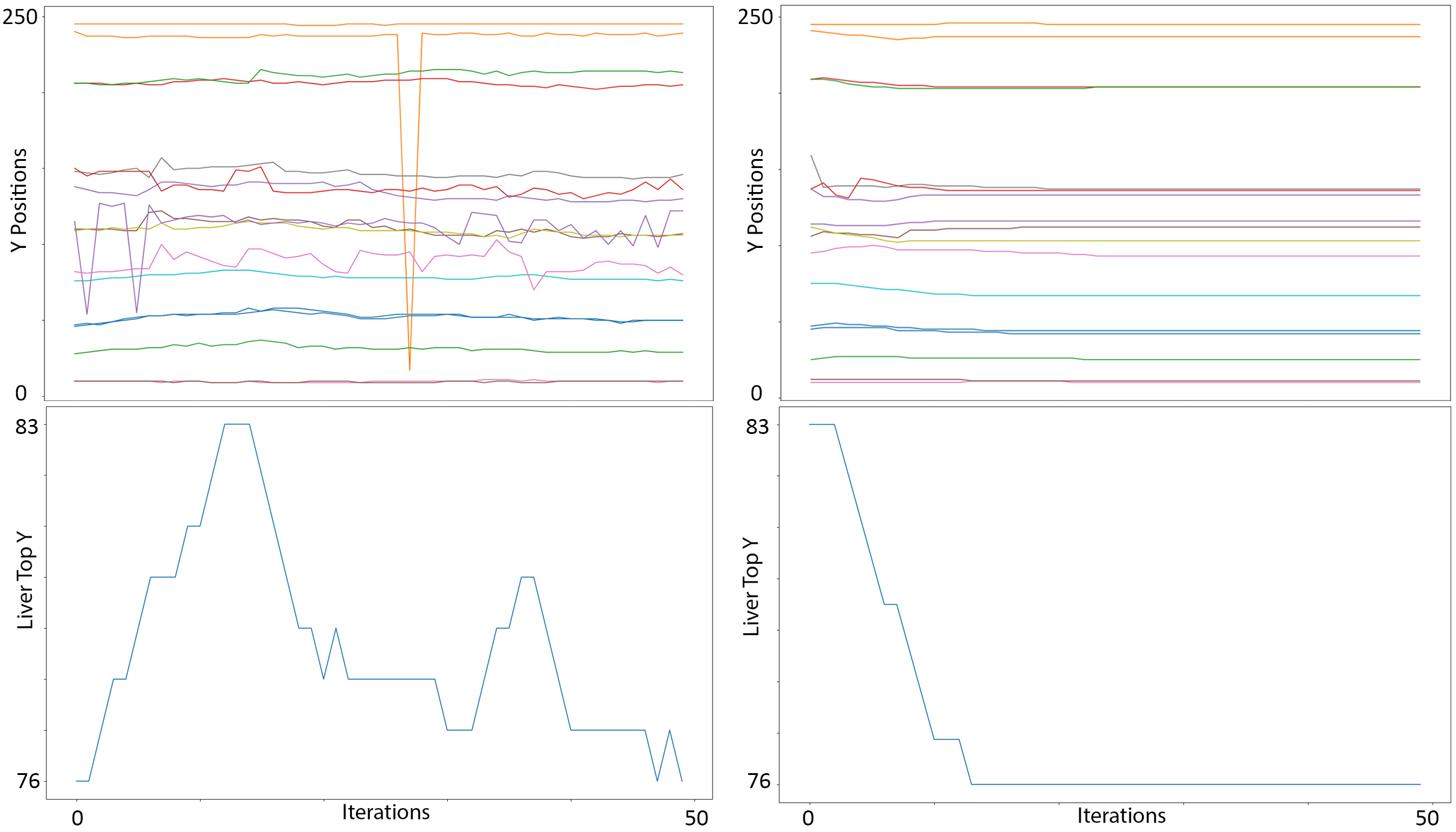}
    \caption{First row shows the evolution of all the 17 markers
      and second row focuses on the `Liver Top' marker, which has the
      largest variation across people. First column presents the results
      before convergence and the convergent results are shown in
      the second column.}
    \label{fig:convergent}
  \end{center}
\end{figure*}

To validate the convergent training, we selected a random data sample
from the testing set and monitored the marker displacement across
iterations. Without the convergent training, the markers kept changing
across iterations. Figure \ref{fig:convergent} shows the variation of
the y position of body markers over 50 iterations. Notice that with the
convergent training, markers become stable after 15 iterations.

\section{Experiments}
To evaluate our approach, we collected $2045$ full body Computed
Tomography (CT) images from patients at several different hospital
sites in North America and Europe. We randomly split the entire dataset into a
testing set of $195$ images, a validation set of $108$ images and a training
set with the rest. The 3D body surface meshes were obtained from the CT
scans using thresholding and morphological image operations. The X-ray
images were generated by orthographically projecting the CT
images. All the data were normalized to a single scale using the neck
and pubic symphysis body markers (since these can be easily
approximated from the body surface data). All the experiments are
conducted in PyTorch environment \cite{pytorch}. Our U-Net like networks
are composed of four levels of convolution blocks (each consisting a 3
repetitions of Convolution, Batch Normalization \cite{batchnorm} and
ReLU \cite{relu}). Each network has 27 convolutional layers, with 32 filters in each layer.

\subsection{Landmark Estimation}
Although the purpose of the convergent training is to ensure a tight
correlation between X-ray and markers, we also computed the error
statistics w.r.t. the ground truth marker annotations provided by medical
experts. Interestingly enough, the convergent training helped improve
the accuracy of the marker prediction network, though the improvement
is not quantitatively significant (the mean euclidean distance dropped
from 2.50 cm to 2.44 cm). An interesting observation is that
the accuracy improved for some of the particularly difficult to
predict markers, such as the tip of the sternum; we attribute this to
the fact that convergent training facilitates more stable
prediction. For example, in case of the tip of the sternum, the error
dropped from 2.46 to 2.00 cm.

\subsection{Synthetic X-ray Prediction}
We use the widely used pix2pix \cite{pixpix} approach as our baseline
for conditional image regression. Following \cite{pixpix}, we use L1-loss for training generators. Using receptive field of $34*34$ achieved the lowest validation error. During training, we found it difficult to optimize the network with batch size $32$. We alleviated this issue by gradually reducing
the batch size from $32$ to $1$; this enabled faster training in beginning although with blurred results but as the training continued and the batch size was reduced to 1, more details were recovered.

\begin{figure*}
	\centering
        \includegraphics[width=0.9\linewidth]{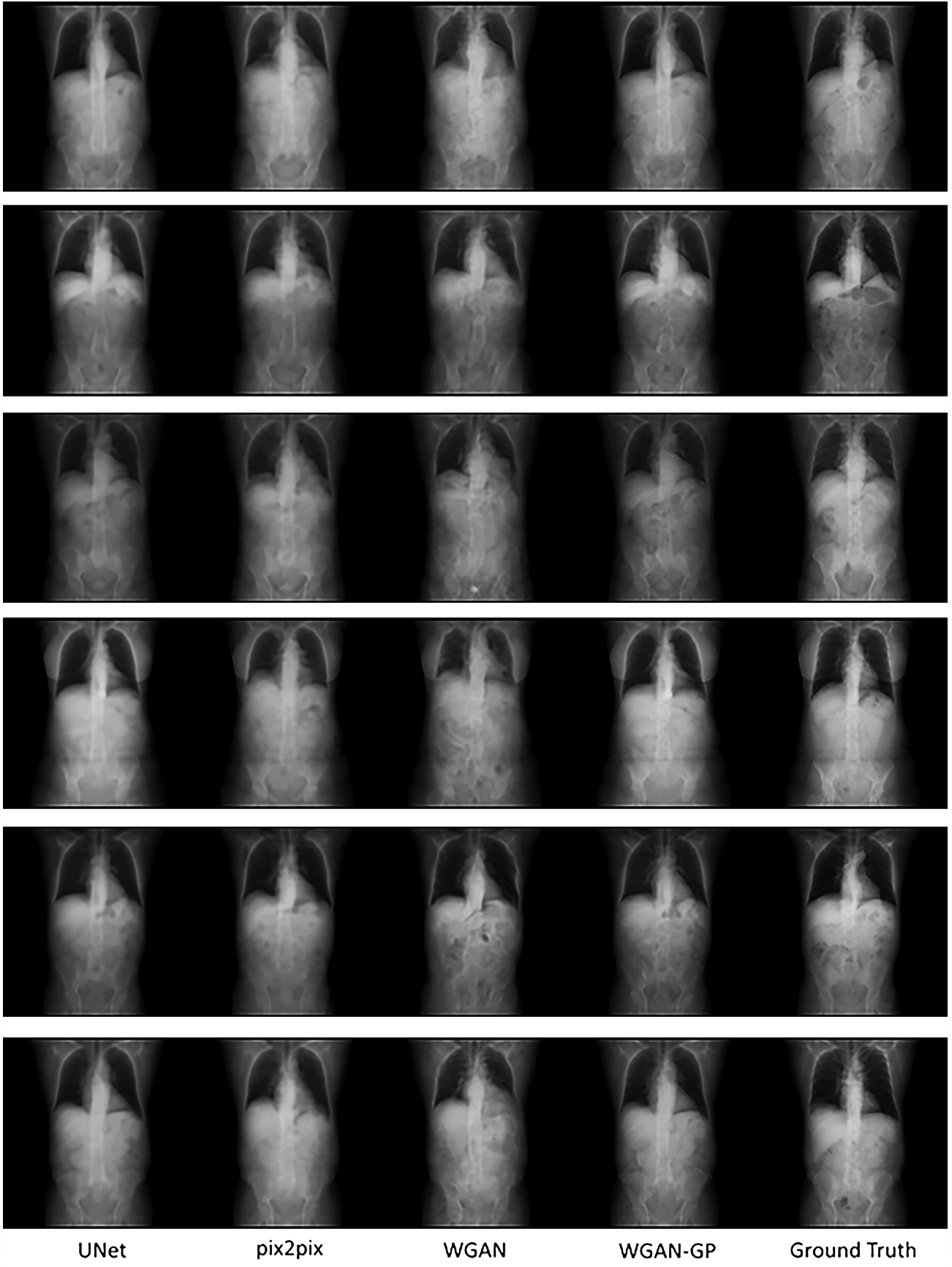}
	\caption{Comparison between X-ray images generated from surface data using various methods (best viewed on screen). The images has been contrast normalized to better appreciate the details in organ boundaries and bone structures. Notice that the images generated using the proposed method are much sharper and even the bone separation in the spine area becomes visible.}
	\label{fig:topogram_results}
\end{figure*}

Table \ref{table:toporesult} shows a quantitative comparison between different approaches. Since L1 error is known to be insufficient in capturing the perceived image quality, we also report the Multi-Scale Structural Similarity index (MS-SSIM) \cite{msssim} averaged over the entire testing set. Notice that the MS-SSIM score for the WGAN-GP is significantly higher than other methods. Furthermore, the  convergent training is able to retain a high perceived image quality, while ensuring that the markers are strongly correlated with the X-ray image. Figure \ref{fig:topogram_results} shows predicted X-ray images for several different surface images using these methods. Compared to ground truth, synthetic X-ray images does surprising well in certain regions such as upper thorax while does poorly in other regions such as lower abdomen where the variance is known to be significantly higher. More importantly, notice the images generated using the proposed method is sharper around organ contours as well as spine and pelvic bone structures are much clearer.

\begin{center}
  \begin{table}
    \resizebox{\columnwidth}{!}{%
      \begin{tabular}{|l|c|c|}
        \hline
        & L1/MAE & Multi Scale SSIM \\
        \hline
        UNet \cite{unet} & $1.62 . 10^{-2}$ & 90.54\\
        \hline
        Pix2Pix \cite{pixpix} & $1.74 . 10^{-2}$ & 90.58\\
        \hline
        WGAN \cite{wgan} & $1.83 . 10^{-2}$ & 89.04\\
        \hline
        WGAN-GP \cite{iwgan} & $1.651 . 10^{-2}$ & 93.96\\
        \hline
	Conv-WGAN-GP & $1.655 . 10^{-2}$ & 93.74\\
        \hline
      \end{tabular}
    }
    \caption{Comparison between state-of-art algorithms on conditional
      image generation/X-ray prediction.}
    \label{table:toporesult}
  \end{table}
\end{center}

\section{Applications}

\subsection{Training Data Generation}
Due to privacy and health safety issues, medical imaging data is
difficult to obtain, which creates a significant barrier for data
driven analytics such as deep learning. Recently, generative deep
learning model such as GANs and other variants \cite{vae, vae_gan,
  cvae, cvae_gan} have been effectively employed to generate realistic
training data, however they offer limited control (categorical) over
the nature of the perturbations.

Parametrized X-ray images offers an approach to generate medical image
training data; furthermore, the spatial parametrization offers
controlled perturbations such as generating data variations with lungs
of certain size. For tasks such as marker detection, we further
observe that since the image manifold is smooth, it's possible to
generate training data (for augmentation) together with annotations,
by annotating the marker in one image and tracking it in the
image domain as it is perturbed along the image manifold.

\begin{figure}[t]
	\includegraphics[width=\linewidth]{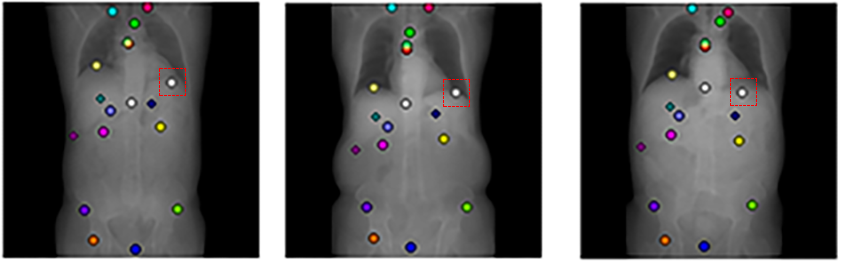}
	\caption{Training examples generated using parametrized
          images from different surface images. The annotated landmark [Left Lung Bottom] is
          surrounded by the dotted box.}
	\label{fig:llb_gen}
\end{figure}

We demonstrate the effectiveness of this method on the task of
detecting left lung bottom landmark (NOTE: this marker is not in the
original set of $17$ landmarks). We manually annotated this marker in
50 synthetic X-ray images to be used for training data. For
evaluation, we manually annotated 50 ground truth X-ray images. To
generate the augmented training dataset, we generate $100$ random
perturbations from the annotated parametrized images (by allowing the
marker to move within a certain range).  Since the image manifold is
smooth, as the position of the marker changed in the perturbed image,
we were able to propagate the annotation using coefficient normed
template matching. Figure \ref{fig:llb_gen} shows sample synthetic
training images.

We train a Fully Convolutional Network \cite{fcn} to regress the
marker location, depicted as a Gaussian mask, from X-ray image. We
used the Adam optimizer with an initial learning rate of $10^{-3}$.

To measure the usefulness of data generated using parametrized images,
we created a baseline training dataset by augmenting the $50$ training
images using $100$ random translations. Table \ref{table:gen_train}
lists the error metrics as the networks are trained using the datasets
for 25 epochs (after which they both overfitted). Notice that after
only 5 epochs, the model trained with the parametrized training had a
$0.99$ cm mean error on the testing set, compared to $8.75$cm for the
baseline. After 25 epochs, the baseline has a mean error of 1.20 cm,
while the network trained on data with parametrized perturbations has
a much lower a 0.68 cm error.

\begin{center}
\begin{table}
  \resizebox{\columnwidth}{!}{%
	\begin{tabular}{|l|l|l|l|l|}
	  \hline Epoch & MSE (p) & MSE (t) &
          $\Delta (p)$ & $\Delta (t)$ \\
	  \hline 1 &
	  $\boldsymbol{6.64 . 10^{-4}}$ & $2.47 . 10^{-3}$ &
	  $\boldsymbol{4.53}$ & $9.17$ \\
	  \hline 5 &
	  $\boldsymbol{4.04 . 10^{-4}}$ & $6.76 . 10^{-4}$ &
	  $\boldsymbol{0.99}$ & $8.75$ \\
	  \hline 10 &
	  $\boldsymbol{3.50 . 10^{-4}}$ & $5.66 . 10^{-4}$ &
	  $\boldsymbol{0.93}$ & $2.67$ \\
	  \hline 15 &
	  $\boldsymbol{3.47 . 10^{-4}}$ & $5.04 . 10^{-4}$ &
	  $\boldsymbol{0.64}$ & $1.97$ \\
	  \hline 20 &
	  $\boldsymbol{3.59 . 10^{-4}}$ & $4.78 . 10^{-4}$ &
	  $\boldsymbol{0.74}$ & $1.28$ \\
	  \hline 25 &
	  $\boldsymbol{3.50 . 10^{-4}}$ & $5.34 . 10^{-4}$ &
	  $\boldsymbol{0.68}$ & $1.20$ \\
	  \hline
	\end{tabular}
  }
  \caption{Comparing translation training set (t) and parametrized
	training set (p). Metrics used are the MSE on the testing set
        and $\Delta$ represents the mean euclidean distance in cm.}
  \label{table:gen_train}
\end{table}
\end{center}

\subsection{X-ray Image Completion}
As the radiation exposure is considered harmful, X-ray images are
often acquired with a limited field of view, only covering a certain
body region (say thorax or abdomen). Using parametric images, we can
reconstruct the X-ray image of the entire body such that it is still
consistent with the partial yet real image. While the reconstructed
X-ray image would be of limited diagnostic use, it would be
beneficial for acquisition planning in subsequent or future medical
scans. Using the reconstructed X-ray, the scan region can be specified
more precisely, thus potentially reducing the radiation exposure.

\begin{figure*}[ht]
  \begin{center}
	\includegraphics[width=\linewidth]{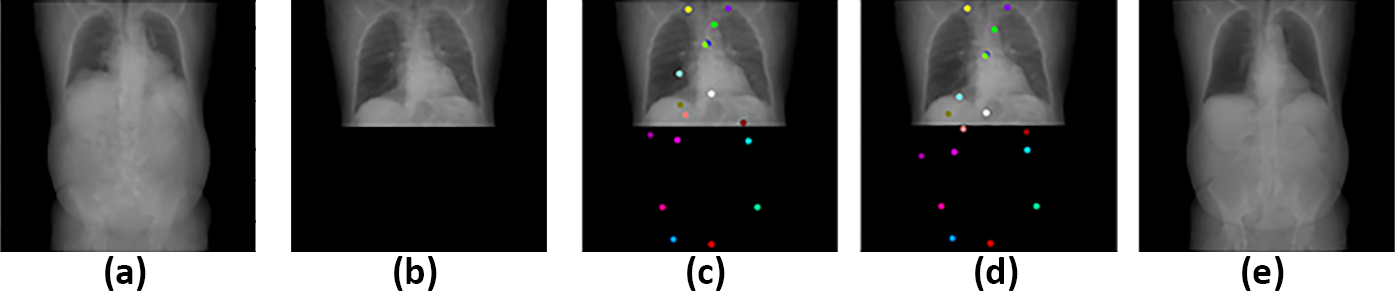}
	\caption{Synthetic X-ray completion to extrapolate an partial X-ray. (a) Predicted X-ray from surface image (b) Partial ground truth X-ray (c) Predicted markers overlaid on ground truth X-ray (d) Adjusted markers using ground truth X-ray (e) Complete synthetic X-ray that matches the ground truth X-ray where data is available.}
	\label{fig:topo_compl}
  \end{center}
\end{figure*}

To reconstruct the complete X-ray, we first generate a parametrized
X-ray image of the patient from the surface data. As previously
mentioned, the predicted X-ray may not always correspond to the true
internal anatomy. This, however, can be addressed using the markers on
the parametrized image by adjusting them such that synthetic X-ray
matches the real one where they overlap. Once the markers are
adjusted, we regenerate the complete X-ray together with all the
markers (see Figure \ref{fig:topo_compl}). Table
\ref{table:completion} shows the quantitative comparison between the
predicted synthetic X-ray and markers, before and after being refined
using the real X-ray image.

\begin{center}
  \begin{table}[h]
      \resizebox{\columnwidth}{!}{%
        \begin{tabular}{|l|l|l|l|l|}
          \hline & MAE & MSE & $\Delta$ & $\Delta_y$ \\
          \hline Initial & $1.77 . 10^{-2}$ & $1.51 . 10^{-3}$ &
          $2.63$ & $2.03$ \\
          \hline Completed & $\boldsymbol{1.41 . 10^{-2}}$ &
          $\boldsymbol{6.60 . 10^{-4}}$ & $\boldsymbol{1.46}$ &
          $\boldsymbol{1.07}$ \\
          \hline
        \end{tabular}
      }
      \caption{Using X-ray completion helped getting a more
        accurate X-ray and significantly reduced the landmarks
        error. $\Delta$ represents the mean euclidean distance and
        $\Delta_y$ the mean y distance. Errors are in cm.}
      \label{table:completion}
    \end{table}
  \end{center}

\subsection{Anomaly Detection}
\begin{figure}
  \centering
  \includegraphics[width=\linewidth]{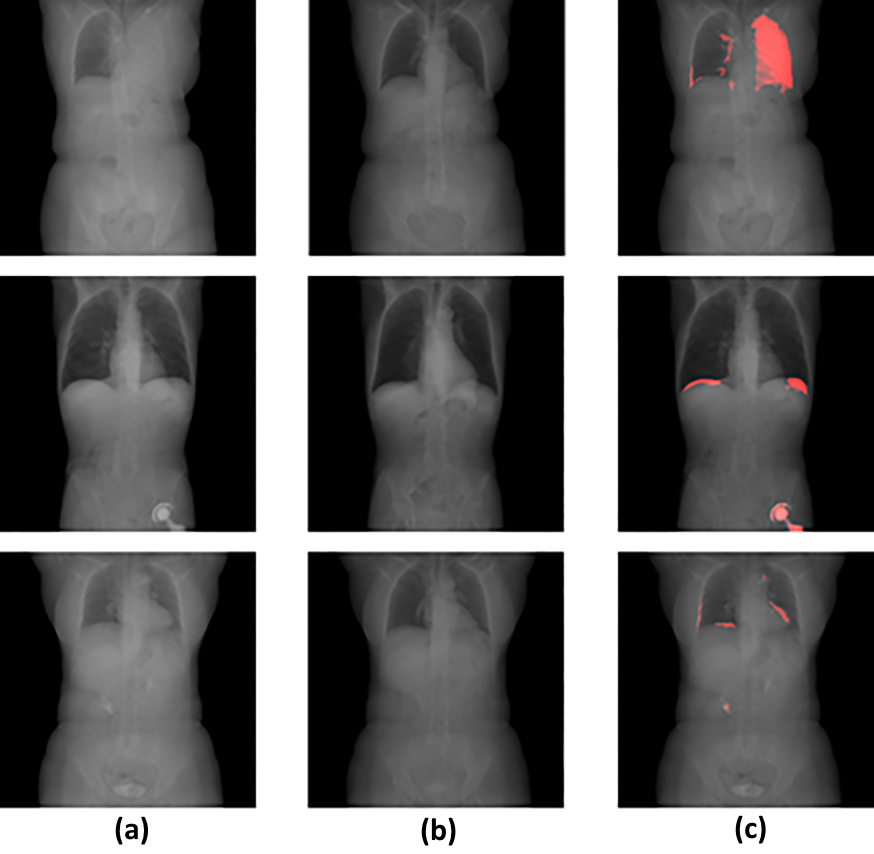}
  \caption{Using the difference between the real X-ray (a) and
    the prediction (b), the anomaly can be easily detected in
    the resulting image (c). Here, the difference threshold was
    set to 10\%. First row shows a missing lung detection, second row
    an implant detection, and last row shows the difference on a healthy patient.}
  \label{fig:anomaly}
\end{figure}

Another potential use case of the proposed method is anatomical
anomaly detection. As the proposed method generates a representation
of healthy anatomy learned from healthy patients, it can be applied
for anomaly detection by quantifying the difference between the real
X-ray image and the predicted one. Figure \ref{fig:anomaly}
illustrates such examples, where one patient has a missing lung, and
the other has implant, which is sometimes overlooked by
technicians. While the anatomical anomaly is easier to identify, the
proposed approach with higher resolution imaging can potentially be
used to suggest candidates for lung nodules (in a chest X-ray) or
other pathological conditions.

\section{Conclusion}
We presented a novel framework that learns to predict parametrized
images from partial image data, which enables natural perturbations of
the predicted images. We apply the presented method on the challenging
task of predicting a synthetic X-ray image from the patient surface data,
together with corresponding markers distributed over the X-ray; the
predicted image can be further manipulated by adjusting the body
markers while ensuring physically consistent X-ray image. The
proposed technology has been demonstrated to address the significant
barrier of training data scarcity in the medical domain, in addition
to enabling novel use cases with benefits to the medical community
such as device positioning. One of our future works is to directly
predict 3D CT from the surface. This work is currently limited by the
difficulties in getting sufficiently large dataset to learn
the variation in 3D anatomical structure. This is also due to the fact that
most of the CT scans are with limited field of view and sometimes may
even have contrast depending on the scan protocol.
{\small \bibliographystyle{ieee} \bibliography{parametrized_images} }

\end{document}